\documentclass[10pt,twocolumn,letterpaper]{article}

\usepackage{cvpr}              

\usepackage{graphicx}
\usepackage{amsmath}
\usepackage{amssymb}
\usepackage{booktabs}
\usepackage{amsfonts}
\usepackage{multirow}
\usepackage{colortbl}
\usepackage{diagbox}
\usepackage{appendix}
\usepackage[accsupp]{axessibility} 
\usepackage[linesnumbered,ruled,vlined]{algorithm2e}
\definecolor{mygray}{gray}{0.91}

\usepackage[pagebackref,breaklinks,colorlinks]{hyperref}

\hypersetup{
	colorlinks=true,
	linkcolor=blue,
	filecolor=blue,      
	urlcolor=blue,
	citecolor=green,
}

\usepackage[capitalize]{cleveref}
\crefname{section}{Sec.}{Secs.}
\Crefname{section}{Section}{Sections}
\Crefname{table}{Table}{Tables}
\crefname{table}{Tab.}{Tabs.}

\def\cvprPaperID{5708} 
\def\confName{CVPR}
\def\confYear{2023}

\newcommand*{\affaddr}[1]{#1} 
\newcommand*{\affmark}[1][*]{\textsuperscript{#1}}

\begin{document}

\title{CASP-Net: Rethinking Video Saliency Prediction from an Audio-Visual Consistency Perceptual Perspective }


\author{
	Junwen Xiong\affmark[1]$^{*}$, 
	Ganglai Wang\affmark[1]$^{*}$ ,  
	Peng Zhang\affmark[1,2]\thanks{Indicates equal contributions.} \thanks{ Corresponding author: zh0036ng@nwpu.edu.cn.} , 
	Wei Huang\affmark[3], 
	Yufei Zha\affmark[1,2], 
	Guangtao Zhai\affmark[4] \\
	\affaddr{\affmark[1]Northwestern Polytechnical University \ \ \ \affmark[2]Ningbo Institute of Northwestern Polytechnical University}\\
	\affaddr{\affmark[3]Nanchang University \ \ \  \affmark[4]Shanghai Jiao Tong University} \\
}

\maketitle


\begin{abstract}

Incorporating the audio stream enables Video Saliency Prediction (VSP) to imitate the selective attention mechanism of human brain. By focusing on the benefits of joint auditory and visual information, most VSP methods are capable of exploiting semantic  correlation between vision and audio modalities but ignoring the negative effects due to the temporal inconsistency of audio-visual intrinsics. Inspired by the biological inconsistency-correction within multi-sensory information, in this study, a consistency-aware audio-visual saliency prediction network (CASP-Net) is proposed, which takes a comprehensive consideration of the audio-visual semantic interaction and consistent perception. In addition a two-stream encoder for elegant association between video frames and corresponding sound source, a novel consistency-aware predictive coding is also designed to improve the consistency within audio and visual representations iteratively. To further aggregate the multi-scale audio-visual information, a saliency decoder is introduced for the final saliency map generation. Substantial experiments demonstrate that the proposed CASP-Net outperforms the other state-of-the-art methods on six challenging audio-visual eye-tracking datasets. For a demo of our system please see our \href{https://woshihaozhu.github.io/CASP-Net/}{project webpage}.

\end{abstract}

\vspace{-15pt}
\section{Introduction}
\label{sec:intro}
The task of saliency prediction is to automatically estimate the most prominent area in a scenario by simulating human selective attention. It has been extended to an alternative way to extract the most valuable information from a massive of data, which serves wide applications such as robotic camera control \cite{butko2008visual}, video captioning \cite{nguyen2013static}, motion tracking \cite{mavani2017facial}, image quality evaluation \cite{zhu2021saliency} and video compression \cite{zhu2018spatiotemporal}, \etc.

\begin{figure}
	\includegraphics[scale=0.23]{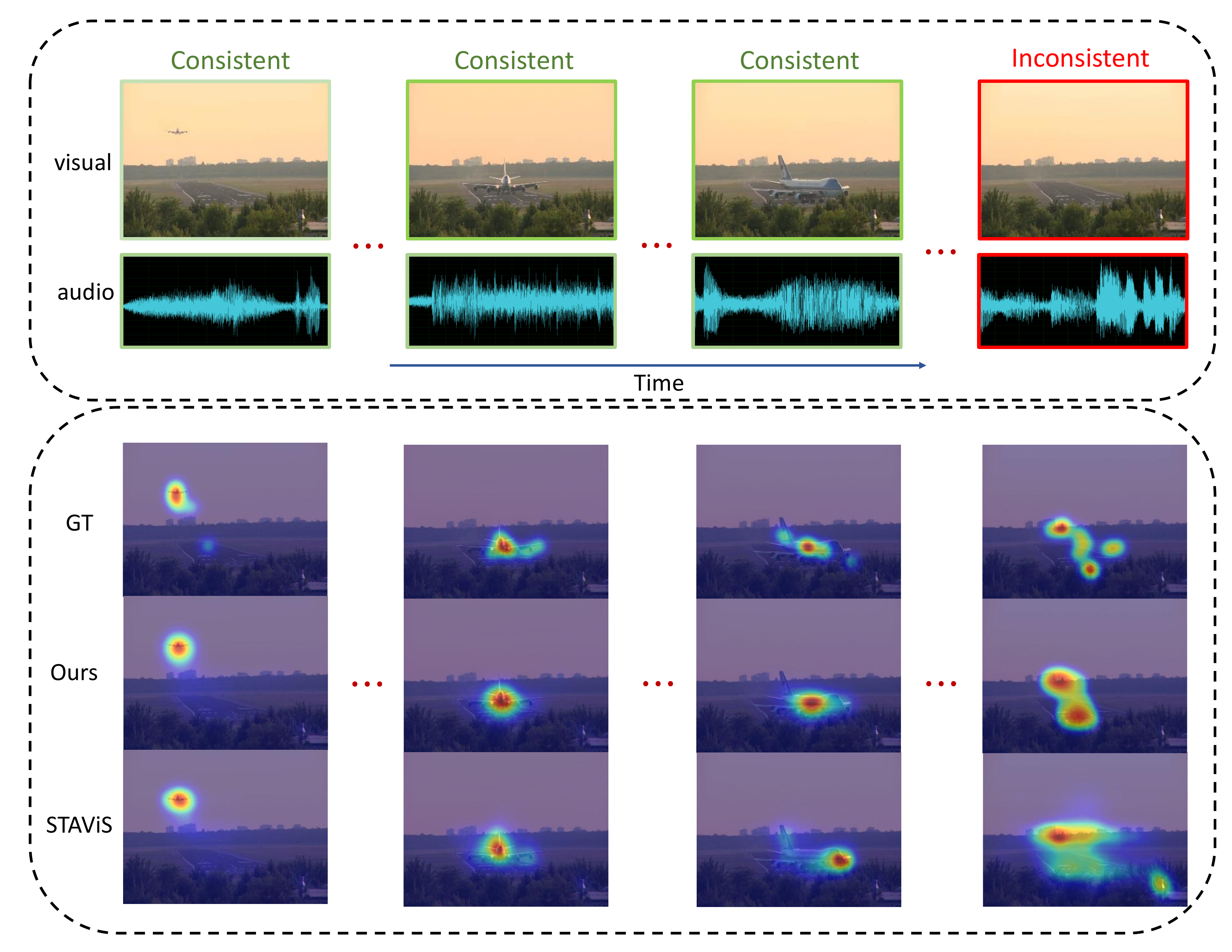}
	\vspace{-20pt}
	\caption{The example figure shows the saliency results of our model compared to STAViS \cite{tsiami2020stavis} in audio and video temporal sequences. In the last time segment, the audio information that occurs in the event is inconsistent with the visual information. Our method can cope with such challenge by automatically learning to align the audio-visual features. The results of STAViS, however, show that it is incapable to address the problem of audio-visual inconsistency. \textit{GT} denotes ground truth.}
	\label{fig-avi}
	\vspace{-10pt}
\end{figure}

In recent years, a lot of saliency prediction works have been developed by their increasing attention \cite{bak2018spatio, wang2018revisiting, cong2018review, vig2014large, wang2017deep,zhang2020spatial}. According to different data types, these studies can be categorized into Image Saliency Prediction (ISP) and Video Saliency Prediction (VSP). The ISP investigates how to combine the low-level heuristic characteristics (\eg, colour, texture and luminance) with high-level semantic image attributes to predict prominent areas in the scene \cite{cong2018review,vig2014large,wang2017deep}. Differently, VSP exploits how to apply the spatio-temporal structure information in videos, and benefits the perception and identification of dynamic scenes \cite{zhang2020spatial,bak2018spatio}.


From the view of data modalities, the vision and audio present the video content from different sensing, which complement each other to enhance the perception. Based on multi-modal data, more recent studies have found that audio information can significantly improve the understanding of the video semantics \cite{min2020multimodal, tsiami2020stavis, tavakoli2019dave}. Min \etal \cite{min2020multimodal} conduct a cross-modal kernel canonical correlation analysis (CCA) by exploring audio-visual correspondence clues, and effectively enhance the video-level saliency prediction accuracy. Tsiami  \etal \cite{tsiami2020stavis} propose a deep model by combining spatio-temporal visual and auditory information to address the video saliency estimation efficiently. 
Nevertheless, these works heavily depend on temporal consistency of visual and audio information, and thus may suffer an unexpected degradation in practical scenarios, where such consistency cannot be satisfied as shown in Figure \ref{fig-avi}.

Temporal inconsistency commonly exists in real-life videos because realistic visual scenarios usually contain multiple sound sources, which may come from on-screen (\eg, dialogue in a talk show), or from off-screen (\eg, narration in a movie). Without understanding the complex scenario components, simply performing audio-visual consistency learning would result in an irrelevant semantic matching. A promising solution to this challenge is motivated by the study of neuroscience \cite{rao1999predictive, friston2005theory}, which explains how our brain minimizes the matching errors within multisensory data using both iterative inference and learning, and also inspired the \textbf{C}onsistency-aware \textbf{A}udio-visual \textbf{S}aliency \textbf{P}rediction network \textit{CASP-Net} of this study.


%
%

By substantially exploring the latent semantic correlations of cross-modal signals, in \textit{CASP-Net}, the potential temporal inconsistency between different modalities can be corrected as well. In addition, a two-stream network is also introduced to elegantly associate video frames with the corresponding sound source, which is able to achieve semantic similarities between audio and visual features by cross-modal interaction. To further reason the coherent visual and audio content in an iterative feedback manner, a consistency-aware predictive coding (CPC) module is designed. Subsequently, a saliency decoder (SalDecoder) is proposed to aggregate the multi-scale audio-visual information from all previous decoder's blocks and to generate the final saliency map. The main contributions in this work can be summarized as follows:


(1) A novel audio-visual saliency prediction model is proposed by comprehensively considering the functionalities of audio-visual semantic interaction  and consistent perception.
(2) A consistency-aware predictive coding module is designed to improve the consistency within audio and visual representations iteratively.
(3) Solid experiments have been conducted on six audio-visual eye-tracking datasets, which demonstrate a superior performance of the proposed method in comparison to the other state-of-the-art works.

\vspace{-5pt}
\section{Related Work}
\label{sec:related}

\subsection{Video Saliency Prediction}
\vspace{-5pt}
For video saliency prediction, different strategies of modeling temporal motion information have been proposed to estimate the saliency maps over consecutive frames \cite{bak2018spatio, min2019tased, jain2021vinet}. Bak \etal \cite{bak2018spatio} propose a two-stream spatio-temporal network to process video frames with optical flow maps, simultaneously. The Long-Short Term Memory (LSTM) \cite{hochreiter1997long}  and Gated Recurrent Unit (GRU) \cite{chung2014empirical} have also been incorporated into the video saliency prediction, and subsequently Wang \etal \cite{wang2018revisiting} propose to combine the Conv-LSTM with dynamic attention mechanism into a network to further enhance the prediction performance. Similarly to model long-term temporal characteristics, Lai \etal \cite{lai2019video} propose STRA-Net based on a lightweight convGRU. After Min \etal \cite{min2019tased} adopt a S3D model \cite{xie2018rethinking} to build TASED-Net with 3D convolutions, the training paradigm of 3D convolution has been widely used in the VSP task. Bellitto \etal \cite{bellitto2020video} design a 3D fully convolutional architecture to obtain multi-scale saliency instances for the combination of output saliency maps. Jain \etal  \cite{jain2021vinet}  adopt a 3D encoder-decoder structure in a U-Net-like fashion, this enables the decoding features of various layers to be constantly concatenated with the corresponding feature of an encoder in the temporal dimension. Also benefit from spatio-temporal modeling of 3D convolution but unlike previous works, a novel saliency decoder is designed in our work to perform aggregation of the multi-scale features.

\begin{figure*}[!htbp]
	\centering
	\includegraphics[width=0.85\textwidth]{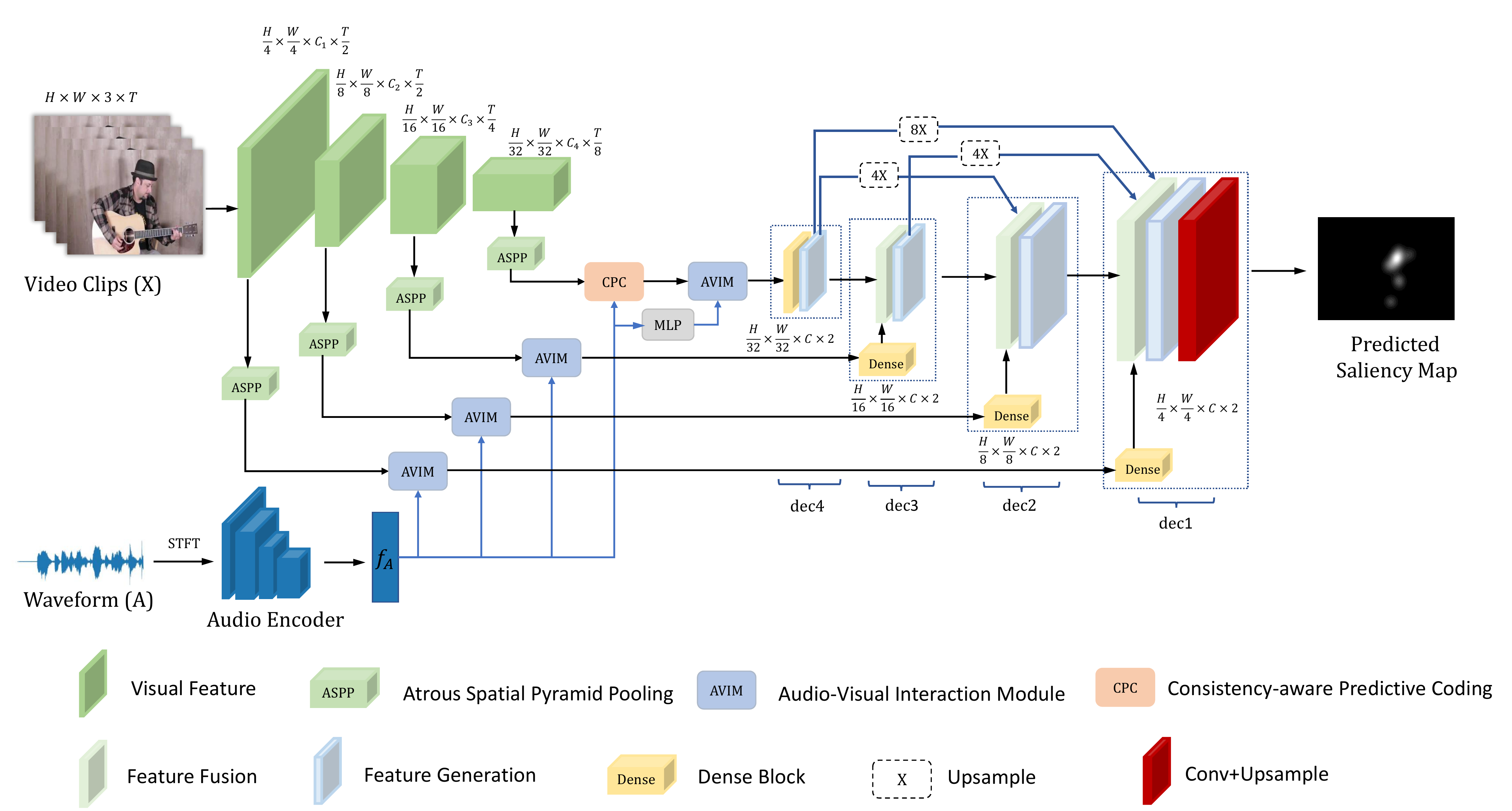}
	\vspace{-10pt}
	\caption{An overview of the proposed CASP-Net. By combining multi-scale visual features and audio features, the designed AVIM and CPC modules enable the network to learn audio-visual semantic interaction and consistent perception. The final saliency decoder utilizes multi-scale audio-visual information to generate saliency maps.
	}
	\label{fig-model_structure}
	 \vspace{-10pt}
\end{figure*}

\subsection{Audio-Visual Saliency Prediction}
\vspace{-5pt}

Early audio-visual saliency prediction methods attempted to establish the cross-modal connections between the two modalities by using CCA \cite{min2016fixation, min2020multimodal}, but an end-to-end deep learning scheme is still far from in-depth study. More recently, Tavakoli \etal \cite{tavakoli2019dave} propose to train two independent 3D ResNet for audio and visual modalities, and the outputs are directly concatenated as a late fusion strategy. With SoundNet \cite{aytar2016soundnet} for audio representation learning, Tsiami \etal\cite{tsiami2020stavis} perform a spatial sound source localization to obtain audio features, which are then fused with the visual feature maps by bilinear operation. In the same way, Jain \etal \cite{jain2021vinet} also employ bilinear fusion operation on the audio features of SoundNet and visual features to predict saliency maps. For these solutions based on a bilinear-based fusion scheme, the large number of learning parameters causes the model learning not easy to converge \cite{tenenbaum2000separating, yu2017multi}. In our work, an attention-based fusion is exploited to learn the cross-modal semantic interaction to overcome such a limitation.



\subsection{Audio-Visual Consistency Learning}
\vspace{-5pt}
For the intrinsic structure of video stream, the audio is naturally paired and synced with the visual component, which means that the audio-visual correspondence can be effectively utilized to draw direct supervision for different tasks: such as visually guided-source separation \cite{gao2021visualvoice}, audio-visual navigation \cite{ chen2021semantic}, active speaker detection \cite{xiong22look}, and audio-visual speech recognition \cite{Afouras18b}, \etc. Unfortunately, most current audio-visual saliency predictions \cite{jain2021vinet, tsiami2020stavis} rely heavily on temporal consistency of visual and audio information,
while ignoring the negative impacts of inconsistent samples.
As a promising solution,  audio-visual consistency detection can be taken into consideration to ensure the performance of saliency prediction \cite{chen2022comprehensive, xuan2020cross}. This also becomes a motivation for this study to devise consistency-aware predictive coding that reasons coherent visual and audio content in an iterative feedback manner.


\section{CASP-Net}
\vspace{-5pt}
\label{sec:Method}
As shown in Figure \ref{fig-model_structure}, the proposed CASP-Net is composed of: a two-stream network to obtain visual saliency and auditory saliency feature, an audio-visual interaction module to integrate the visual and auditory conspicuity maps, a consistency-aware predictive coding module to reason the coherent spatio-temporal visual feature with audio feature, and a saliency decoder to estimate saliency map with multi-scale audio-visual features. Each part is elaborated in the below.


\subsection{Two-Stream Encoders}
\label{sec:Encoder}
\vspace{-5pt}
Let $X \in  \mathbb{R}^{  H_v  \times W_v \times  3 \times  T_v}$ and $A \in \mathbb{R}^{T_A}$ denote video frames and the corresponding audio signal, respectively.

\noindent \textbf{Video Encoder}: We employ the off-the-shelf S3D \cite{xie2018rethinking} as a video backbone network to encode the spatio-temporal information. This is because that S3D is lightweight and pre-trained on a large dataset, which makes it fast and effective for transfer learning. The backbone consists of 4 convolutional stages, and outputs hierarchical visual feature maps during the encoding process, as shown in Figure \ref{fig-model_structure}. The generated features are denoted as $f_{X_i} \in \mathbb{R}^{ h_i  \times w_i \times C_i \times T_i}$, where $(h_i, w_i) = (H_v, W_v)/2^{i+1}, i=1,...,4$.

\noindent \textbf{Audio Encoder}: For audio representation, the 1D audio waveform needs to be converted into the 2D spectrogram by Short-Time Fourier Transform (STFT). Instead of directly applying 1D CNNs on time domain audio signals, a 2D fully convolutional network is employed for this operation.
When audio is cropped to match the visual frames duration (\eg, $T_v=16$), the log-Mel spectrogram is calculated for each matched signal by taking absolute values of a complex STFT, following the natural logarithm. For the high-level semantic information, we employ the VGGish network \cite{hershey2017cnn} with pre-trained weights on AudioSet. An audio embedding is generated as the original audio feature $f_A  \in \mathbb{R}^{C_A}$ from the layers before the final post-processing stage.

\subsection{Cross-Modal Semantic Interaction}
\vspace{-5pt}
To find cross-modal semantics with implicitly associated audio and visual representations in videos, we perform Atrous Spatial Pyramid Pooling (ASPP) \cite{chen2017deeplab} on the post-process of visual features $f_{X_i}$ to $f_{V_i} \in \mathbb{R}^{h_i \times w_i \times C \times T_i}$, where $C=256$. With multiple parallel filters of different rates, the pooling operation helps to recognize visual objects with different receptive fields, \eg, different-sized moving objects.

\vspace{-5pt}
\begin{figure}
	\includegraphics[scale=0.195]{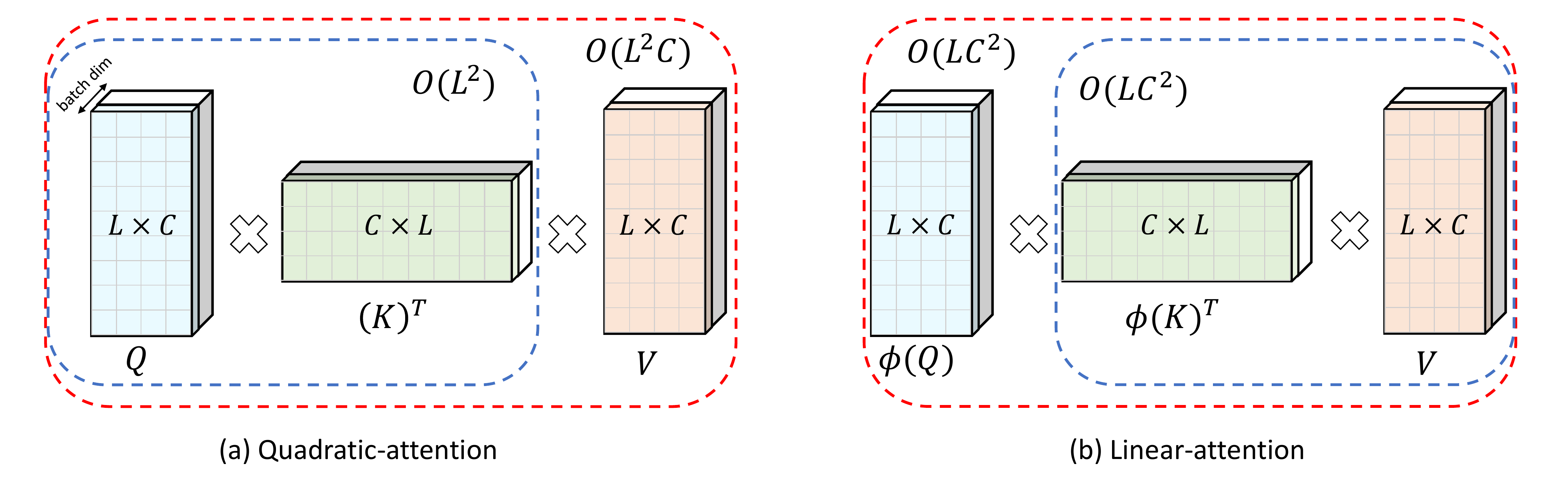}
	\vspace{-20pt}
	\caption{Quadratic $vs.$ Linear. $L$ denotes the number of tokens ($T_{i}h_{i}w_{i}$) in feature, and $C$ denotes a constant (\ie, $C=256$).
		Quadratic-attention scales with the \textit{square} of the $L$. Using a decomposable kernel $\phi(\cdot)$,  we can rearrange the order of
		operations such that linear attention scales \textit{linearly} with $L$. Dashed-blocks indicate order of computation with corresponding time complexities attached.}
	\label{fig-avim}
	\vspace{-15pt}
\end{figure}

Considering visual and auditory features have different feature dimensions, an affine transformation is applied to the audio feature to match the channel of visual feature $f_{V_i}$. Then it is duplicated $h_{i}w_{i}T_i$ times in spatio-temporal dimension, and reshaped to the same size as $f_{V_i}$, which is denoted as $f_{\hat{A}}$. To learn the correspondence between the audio and visual features $f_{\hat{A}}$, $f_{V_i}$, two different audio-visual interaction approaches are investigated:

\noindent \textbf{Audio-Visual Interaction (Quadratic)}: For the encoding of audio-video correlations, the mechanism of self-attention \cite{vaswani2017attention} is adopted. According to the calculation method of self-attention , the audio and visual feature matrices need to be transformed into the vector format as $f_{V_i} \in \mathbb{R}^{T_{i}h_{i}w_{i} \times C}$ and  $f_{\hat{A}} \in \mathbb{R}^{T_{i}h_{i}w_{i} \times C}$, respectively. Such an audio-visual interaction can be measured by dot-product, then the updated feature maps $f_{V_i}$ at the $i$-th stage becomes,
\vspace{-5pt}
	\begin{equation}
		\begin{aligned}
			& Q = \alpha(f_{V_i}); \ K = \beta(f_{\hat{A}}); \ V = \gamma(f_{V_i}) \\
			&\widetilde{A} = softmax(\frac{Q K^T}{N})V  \\
			&f_{V_i} = f_{V_i} + \delta(\widetilde{A})
			\label{eq_soft_attn}
		\end{aligned}
	\end{equation}

\noindent where $\alpha, \beta, \gamma$ and $\delta$ are $1\times1\times1$ convolutions, $N = T_{i} \times h_{i}\times w_{i}$ is a scale factor, and $\widetilde{A}$ denotes the audio-visual similarity matrix. Each visual pixel can be associated with all auditory information by audio-visual interaction.

However, Equation \ref{eq_soft_attn} shows that the computational cost of self-attention increases quadratically with the number of tokens $(T_{i}h_{i}w_{i})$ in feature. The same is true for the memory requirements because the similarity matrix $\widetilde{A}$ must be saved to calculate the gradients with respect to the $Q$, $K$ and $V$ (see also Figure \ref{fig-avim}(a)).

\noindent \textbf{Audio-Visual Interaction (Linear)}: As in \cite{katharopoulos2020transformers}, the audio-visual similarity matrix $\widetilde{A}$ can be generalized by treating $softmax(\cdot)$ as a pairwise similarity between $Q$ and $K$. That is, for some similarity function $sim(\cdot)$, we have,
		\vspace{-5pt}
		\begin{equation}
			\widetilde{A} = sim(Q,K)V
			\vspace{-5pt}
		\end{equation}

If choose a decomposable kernel with feature representation $\phi(\cdot) \ge 0$ as $sim(x,y) = \phi(x)\phi(y)^T$, we have
		\begin{equation}
			\widetilde{A}(\phi) = (\phi(Q)\phi(K)^T)V
			\vspace{-5pt}
		\end{equation}

Then by associativity, the order of computation can be changed as,
		\begin{equation}
			\widetilde{A}(\phi) = \phi(Q)(\phi(K)^TV )
		\end{equation}

\noindent which allows us to compute $\phi(K)^TV$. This leads to an operation $O(T_{i}h_{i}w_{i}C^2)$ to create a $C^2$ matrix instead of a $(T_{i}h_{i}w_{i})^2$ one, where $C$ is usually much less than $T_{i}h_{i}w_{i}$ (see also Figure \ref{fig-avim}(b)). To implement the similarity matrix, the following kernel function is designed as:
\vspace{-5pt}
	\begin{equation}
			\phi(x) = gelu(x) + 0.2
			\vspace{-5pt}
		\end{equation}

\noindent where $gelu(\cdot)$ denotes the gaussian error linear units \cite{hendrycks2016gaussian}.

\begin{figure}
	\includegraphics[scale=0.35]{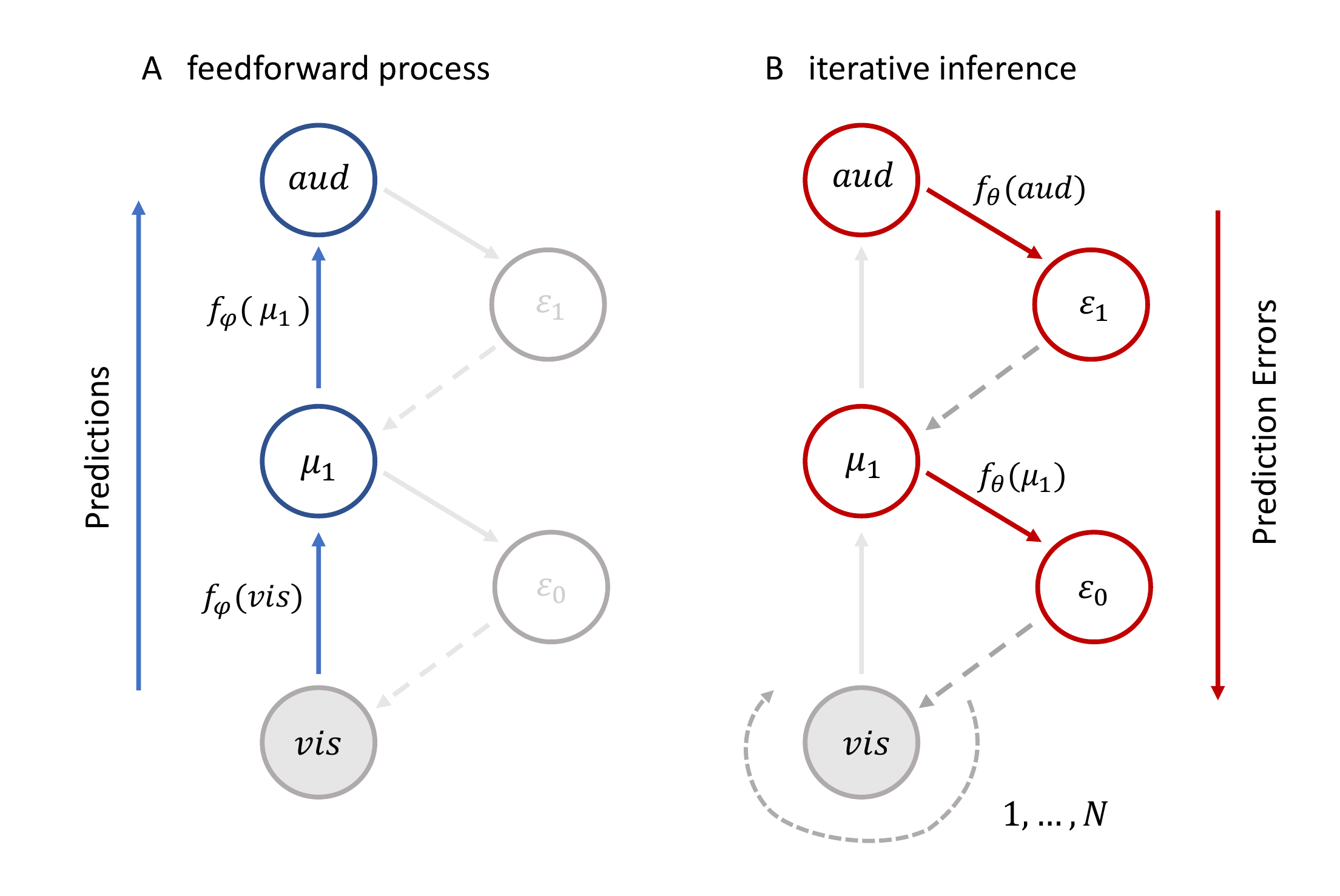}
	\vspace{-15pt}
	\caption{Consistency-aware Predictive Coding combines two different phases as follows. For  brevity, we use $vis$ and $aud$ to denote $f_{V_4}$ and $f_A$, respectively. (A) The $vis$ propagated up the hierarchy in a feedforward manner, utilising the non-linear function $f_{\varphi}(\cdot)$. (B) The initial values for $\mu$ are then used to predict the activity at the layer below, transformed by the iterative functions $f_{\theta}(\cdot)$. These predictions incur prediction error $\epsilon$, which are then used to update activity $\mu$. This process is repeated $N$ times, after which perceptual inference is complete.
	}
	\label{fig-cmpc}
	 \vspace{-10pt}
\end{figure}

\subsection{Consistency-aware Predictive Coding}
\vspace{-5pt}
To overcome the potential inconsistencies introduced by audio and visual features, consistency-aware predictive coding (CPC) is proposed to improve the performance of reasoning multi-modal features. Inspired by the predictive coding (PC) in theoretical neuroscience \cite{buckley2017free, alink2010stimulus}, it represents top-down signalling in the perceptual hierarchy to predict the cause of sensory data. For predictive coding, each hierarchical layer predicts the activity of the layer below it (with the lowest layer predicting the sensory data). Such predictions are then iteratively refined by minimizing the prediction errors, (\ie, the difference between predictions and the actual activity), in each layer \cite{ friston2005theory, rao1999predictive}. Comparatively, the proposed CPC adopts bottom-up $f_{\varphi}(\cdot)$ and top-down  $f_{\theta}(\cdot)$ paths to represent the prediction and iterative process respectively, as shown in Figure \ref{fig-cmpc}. The CPC inputs the visual feature $vis$ as a type of prior knowledge to predict the audio feature $aud$ iteratively. It is composed of $L$ hierarchical layers, in which each layer $i$ is composed of a variable unit $\mu_i$ and an error unit $\epsilon_i$.



\noindent \textbf{Feedforward Process}: A set of feedforward parameters $\varphi$ are defined in correspondence to bottom-up connections. The feedforward parameters represent non-linear functions which map activity at one layer to activity at above layer. Thereby a bottom-up prediction is implemented as:
\vspace{-5pt}
\begin{equation}
\mu_i = f_{\varphi}(\mu_{i-1})
\vspace{-3pt}
\end{equation}

\noindent \textbf{Iterative  Inference}: The iterative phase updates the activity $\mu_i$ again by generating top-down prediction $f_{\theta}(\mu_{i+1})$ with the prediction error $\epsilon_i^{pred}$.
\vspace{-5pt}
\begin{equation}
    \begin{aligned}
	  &\epsilon_{i}^{pred} = MSE(\mu_i - f_{\theta}(\mu_{i+1})) \\
	  & \mu_i \leftarrow  \mu_i - \alpha \nabla \epsilon_{i}^{pred}
	 \end{aligned}
	 \label{eq_error}
	 \vspace{-5pt}
\end{equation}

\noindent where $\nabla \epsilon_{i}^{pred}$ denotes the gradient with respect to the activity of $\mu_{i+1}$, the hyper-parameter $\alpha$ is introduced to reduce the prediction error.

To decrease the prediction error in CPC, the operations of feedforward process and iterative inference are alternately performed while gradually improving the representations of all layers. When the entire inference ends, the bottom layer is directly output with perceptually consistent audio-visual features.


\subsection{SalDecoder}
\vspace{-5pt}

A new decoder architecture is also proposed for saliency estimation. The decoder consists of 4 blocks, \ie, dec$_x$, where $x$ = $1, ..., 4$. Figure \ref{fig-model_structure} presents the overview of the decoder's block. Each of the blocks consists of a dense block  \cite{huang2017densely} and a feature generation block. Moreover, the blocks dec$_1$, dec$_2$, and dec$_3$ have a fusion function among them named fusion blocks, which combine the output of the previous decoder with the output from the dense block. The final saliency map is obtained using the combination of Conv and Upsample in the dec$_1$ stage.

Specifically, the audio-visual feature representations $\{f_{X_i}\} (i=1,...,4)$ are taken as input from AVIM to the decoder. In each stream,  the dense block is firstly utilized to process representation by its feature propagation, and align the temporal dimensions of each feature to facilitate subsequent fusion. All blocks except dec$_4$ contain a fusion block to integrate multi-scale features. For the explicit operation in the fusion block, each feature is initialized to align in the spatial dimension via upsampling, and fused by element-wise summation. The fused feature are fed into the feature generation block, which consists of BN-ReLU-3D Conv layers, to obtain the semantic features with context information. In the dec$_1$, the combination of 3D Conv and Upsample is also performed to generate the final saliency map.


\subsection{Saliency Losses}
\vspace{-5pt}
We refer to the training paradigm of multiple loss functions in \cite{tsiami2020stavis, chang2021temporal}, which contains: Kullback-Leibler ($KL$) divergence, Linear Correlation Coefficient ($CC$) and Similarity Metric ($SIM$). Assuming that the predicted saliency map is $S_{pred} \in [0, 1]$, the labeled binary fixation map is $S_{fix} \in \{0, 1\}$, and the dense saliency map generated by the fixation map is $S_{den} \in [0,1]$, then $L_{KL}$, $L_{CC}$, and $L_{NSS}$ are employed to signify three different loss functions, respectively. The first is the $KL$ loss between the predicted map $S_{pred}$ and the dense map $S_{den}$:

\vspace{-10pt}
\begin{equation}
	L_{KL} (S_{pred}, S_{den}) = \sum_{x}^{}S_{den}(x)ln\frac{S_{den}(x)}{S_{pred}(x)}
\end{equation}
\vspace{-10pt}

\noindent where $x$ represents the spatial domain of a saliency map. The second loss function is based on the $CC$ that has been widely used in saliency evaluation, and used to measure the linear relationship between the predicted saliency map $S_{pred}$ and the dense map $S_{den}$:
\vspace{-10pt}
\begin{equation}
	L_{CC} (S_{pred}, S_{den}) = - \frac{cov(S_{pred}, S_{den})}{\rho(S_{pred})\rho(S_{den})}
\end{equation}

\noindent where $cov(\cdot)$ and $\rho(\cdot)$ represent the covariance and the standard deviation respectively. The last one is derived from the $SIM$, which can measure the similarity between two distributions:
\vspace{-10pt}
\begin{equation}
	\begin{aligned}
		L_{SIM}(S_{pred}, S_{den}) =  \sum_{x}^{} min\{\zeta(S_{pred}(x)), \zeta(S_{den}(x)) \} \\
	\end{aligned}
\end{equation}

\noindent where $\zeta$ represents the normalization operation. The weighted summation of the above $KL$, $CC$ and $SIM$ is taken to represent the final loss function:
\vspace{-5pt}
\begin{equation}
	L_{total} =  L_{KL} + \lambda_1 L_{CC} + \lambda_2 L_{SIM}
	\vspace{-5pt}
\end{equation}

\noindent where $\lambda_1$, $\lambda_2$ are the weights of $CC$ and $SIM$, respectively. 

\section{Experiment}
 \vspace{-5pt}
Experiments are conducted on a total of seven datasets including a pure visual dataset and six audio-visual eye-tracking datasets. In the following subsections, the implementation details and evaluation metrics are firstly introduced. We represent the experimental results with analysis via the ablation studies and comparison with the state-of-the-art works.

\begin{table}[!tbp]
	\resizebox{\linewidth}{!}{
		\begin{tabular}{l|l|lllll}
			\multicolumn{1}{c|}{Data} & \multicolumn{1}{c|}{Method}  & \multicolumn{1}{c}{\#Params} & CC$\uparrow$ & NSS$\uparrow$  & SIM$\uparrow$ \\
			\hline
			\multirow{3}*{AVA} &CASP-Net(FCN) &  \textbf{1.77}  & 0.659  & 3.57& 0.509    \\
			&CASP-Net(UNet) & 3.94 &   0.665   &   3.65   & 0.512   \\
			&\textbf{CASP-Net(Sal)} &  2.49 & \textbf{0.671}     & \textbf{3.67} &  \textbf{0.515}   \\
			\hline
			\multirow{3}*{ETMD}  &CASP-Net(FCN) &  \textbf{1.77}  & 0.606  & 3.21  & 0.463    \\
			&CASP-Net(UNet) & 3.94 &   0.611   &   3.27 & 0.468   \\
			&\textbf{CASP-Net(Sal)} &  2.49 & \textbf{0.613}     & \textbf{3.30} &  \textbf{0.471}   \\
	\end{tabular}}
	\vspace{-5pt}
	\caption{Comparison between SalDecoder and the other different decoders on AVAD and ETMD datasets (visual-only). \#Params represent the number of parameters per decoder.}
	\label{table-ablation_1}
	\vspace{-15pt}
\end{table}

\vspace{-5pt}
\subsection{Setup}
\subsubsection{Datasets}
\vspace{-5pt}
\noindent \textbf{Visual Dataset}: The DHF1k \cite{wang2018revisiting} is one of the most popular visual-only datasets in the study of video saliency, It contains 1000 videos where 600 videos are for training and 100 for validation. In addition, a test set of 300 videos is also released but without public ground truth. Considering that the main focus of our model is on multi-modal scenarios, its visual branch is pre-trained using this dataset.

\begin{table}[!tbp]
	\resizebox{\linewidth}{!}{
		\begin{tabular}{lllll}
			\toprule
			\multirow{2}{*}{Method} & \multicolumn{2}{c}{AVAD}  & \multicolumn{2}{c}{ETMD}  \\
			\cmidrule(r){2-3}  \cmidrule(r){4-5}
			& CC $\uparrow$           & SIM   $\uparrow$ & CC	$\uparrow$	  & SIM $\uparrow$ \\
			\midrule\midrule
			Visual-Only                                  & 0.671 & 0.515 & 0.613  & 0.469     \\
			\hline
			V+A+Bilinear                                   & 0.670 & 0.510 & 0.609  & 0.469    \\
			
			V+A+I$_{Quadratic}$                             & 0.674 & 0.517 & 0.615  & 0.471    \\
			
			V+A+I$_{Linear}$   						       & 0.675 & 0.519 & 0.615  & 0.470   \\
			V+A+I$_{Linear}$ +C                            & \textbf{0.685} & \textbf{0.528} & \textbf{0.616}  & \textbf{0.476}     \\
			\bottomrule
	\end{tabular}}
	\vspace{-5pt}
	\caption{Ablation Studies. The visual-only (V) denotes the visual branch of CASP-Net. A refers to the audio branch, I refers to the audio-visual interaction module, C refers to the consistency-aware predictive coding, and the subscripts represent two schemes with different computational complexity.}
	\label{table-ablation_2}
	\vspace{-10pt}
\end{table}

\begin{table}[!tbp]
	\resizebox{\linewidth}{!}{
		\begin{tabular}{llllll|lll}
			\toprule
			\multirow{2}{*}{Data} & \multirow{2}{*}{Metric} & \multicolumn{6}{c}{$i$-th stage of Video Encoder, $i \in \{1, 2, 3, 4 \}$}  \\
			\cmidrule(r){3-8}  
			& & 1        & 2      & 3	     & 4  & 2,3,4  & 1,2,3,4 \\
			\midrule\midrule
			\multirow{2}*{AVA} & CC     & 0.670 & 0.672 & \textbf{0.673}  & 0.671  & 0.674 & \textbf{0.675}   \\
								& SIM   & 0.510 & 0.512 & \textbf{0.514}  & 0.513 & 0.515 & \textbf{0.518}    \\
			\hline
			\multirow{2}*{ETMD} & CC   & 0.610 & \textbf{0.613} & 0.611  & 0.611  & 0.613 & \textbf{0.615}    \\
							& SIM      & 0.467 & \textbf{0.469} & 0.468  & 0.467  & \textbf{0.471} & 0.470  \\
			\bottomrule
	\end{tabular}}
	\vspace{-5pt}  
	\caption{Audio-visual Interaction at various video encoder stages. In both the AVAD and ETMD datasets, the model achieves almost the best performance when the AVIM is used in all four stages.}
	\label{table-AVIM-ablation}
	\vspace{-15pt}
\end{table}

\noindent \textbf{Audio-Visual Dataset}: There are six audio-visual datasets in video saliency: AVAD \cite{min2016fixation}, Coutrot1 \cite{coutrot2014saliency}, Coutrot2 \cite{coutrot2016multimodal} , DIEM \cite{mital2011clustering}, ETMD \cite{koutras2015perceptually}, and SumMe \cite{gygli2014creating}, which are used for our evaluation comparison. In detail, (i) the AVAD dataset contains 45 video clips with a duration of 5-10 seconds. These clips cover a variety of audio-visual activities, \eg, playing the piano, playing basketball, making an interview, \etc. The dataset  contains eye-tracking data from 16 participants. (ii) The Coutrot1 and Coutrot2 datasets are separated from the Coutrot dataset. The Coutrot1 dataset contains 60 video clips covering 4 visual categories: one moving object, several moving objects, landscapes, and faces. The corresponding eye-tracking data are from 72 participants. The Coutrot2 dataset contains 15 video clips, which record 4 persons having a meeting. The corresponding eye-tracking data are from 40 participants. (iii) The DIEM dataset contains 84 video clips including game trailers, music videos, advertisements and \etc. which are captured from 42 participants. It should be noted that the audio and visual tracks in these videos do not correspond naturally. The ETMD dataset contains 12 video clips from several Hollywood movies, with the eye-tracking data annotated by 10 different people. The SumMe dataset consists of 25 video clips with diverse topics, \eg, playing ball, cooking, travelling, \etc., and the corresponding eye-tracking data are collected from 10 viewers.

\vspace{-10pt}
\subsubsection{Implementation Details}
\vspace{-5pt}
We use pre-trained S3D model \cite{xie2018rethinking} on Kinetics \cite{carreira2017quo} and pre-trained VGGish \cite{hershey2017cnn} on AudioSet \cite{gemmeke2017audio}. The input samples of the network consist of 16-frame video clips of size $224 \times 384 \times 3$ with the corresponding audio stream, which is transformed into $96 \times 64$ log-Mel spectrograms. For a fair comparison,  two different training strategies are designed depending on whether the DHF1k dataset is used or not. The first is to train the entire model on the six audio-visual datasets from scratch. The other is to train the visual branch of the model on the DHF1k dataset, and then use this weight to fine-tune the entire model on these audio-visual datasets. Both strategies end up with the same evaluation as \cite{tsiami2020stavis}.

For the CPC module, we mainly adopt 3D Conv layers in the feedforward process and 3D Deconv layers in the iterative inference. The hyper-parameter $\alpha$ is set to $0.1$. The proposed training process chooses Adam as the optimizer with the started learning rate of 1e-4.   The loss weight is to $-0.1$, \ie, $\lambda_1 = \lambda_2 =-0.1$. The computation platform is configured by an NVIDIA GeForce RTX 3090 GPU with batch-size 8 for entire experiments.


\begin{figure}[!tbp]
	\hspace{-1cm}
	\flushright
	\begin{minipage}[t]{0.49\linewidth}
		\centering
		\includegraphics[width=\linewidth]{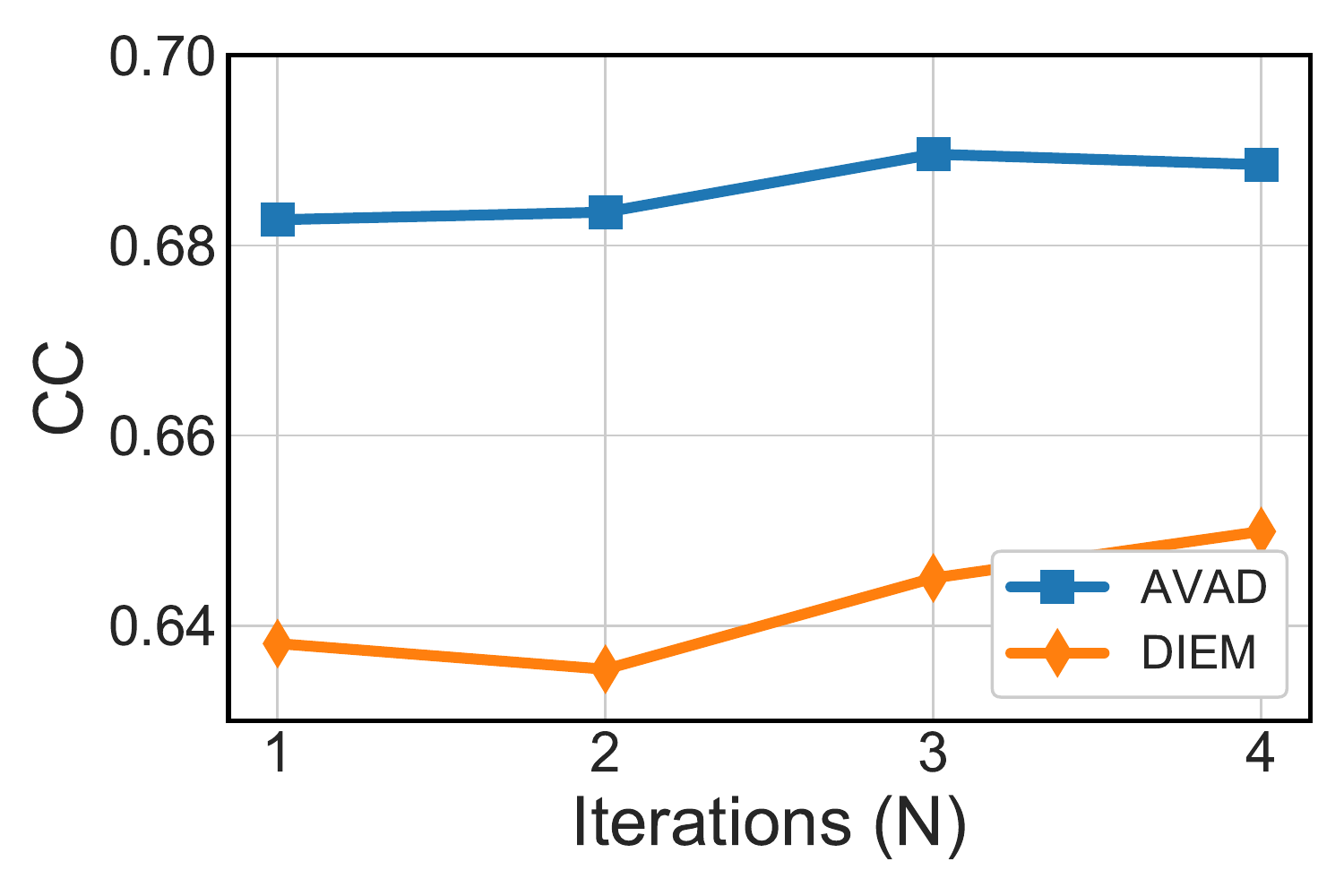}
	\end{minipage}
	\begin{minipage}[t]{0.49\linewidth}
		\centering
		\includegraphics[width=\linewidth]{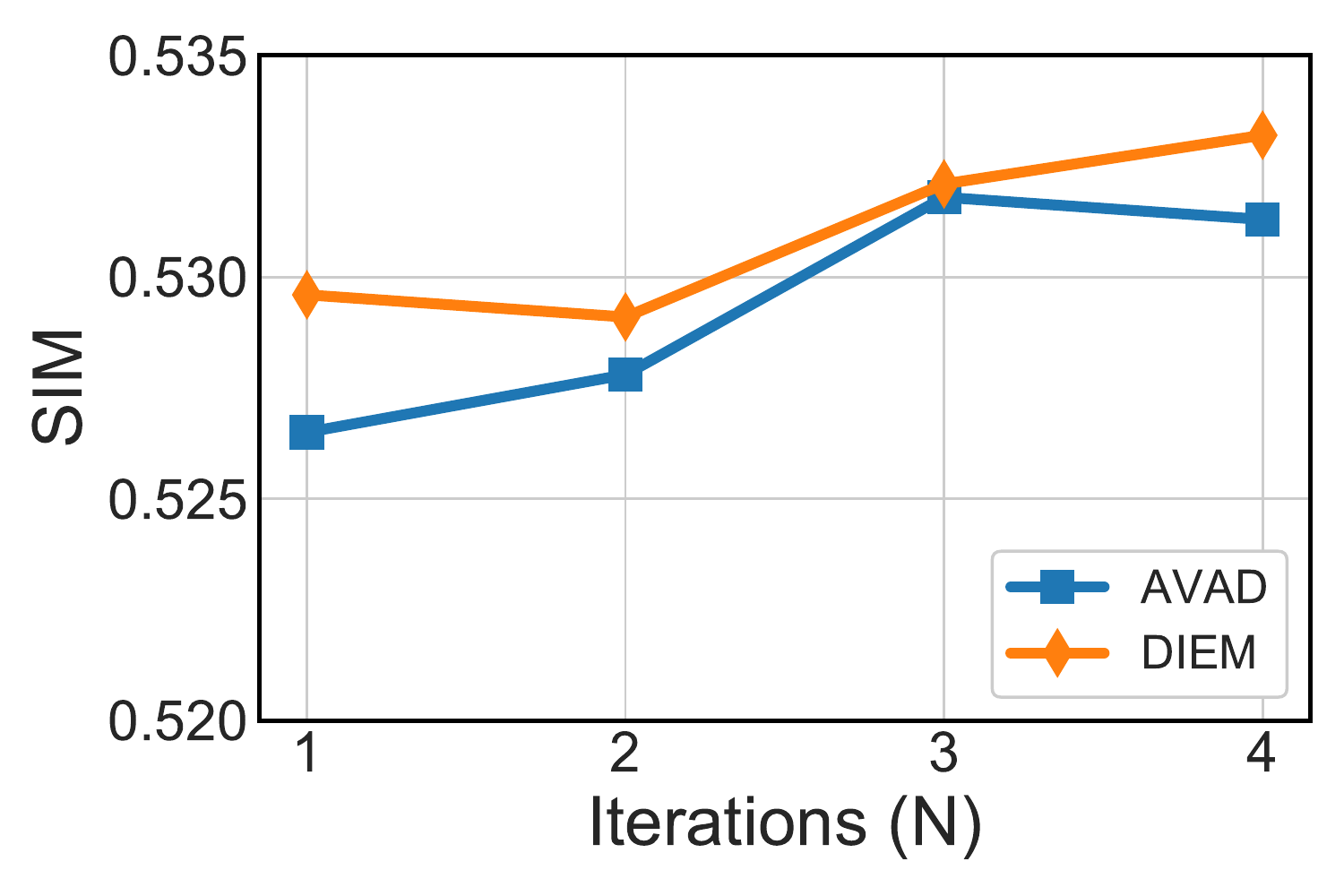}
	\end{minipage}
	\vspace{-10pt}
	\caption{Performance analysis of CPC's iterations on AVAD and DIEM datasets.}
	\label{fig:exp_cpc_iteration}
	\vspace{-10pt}
\end{figure}

\begin{figure}[!tbp]
	\flushleft
	\includegraphics[scale=0.53]{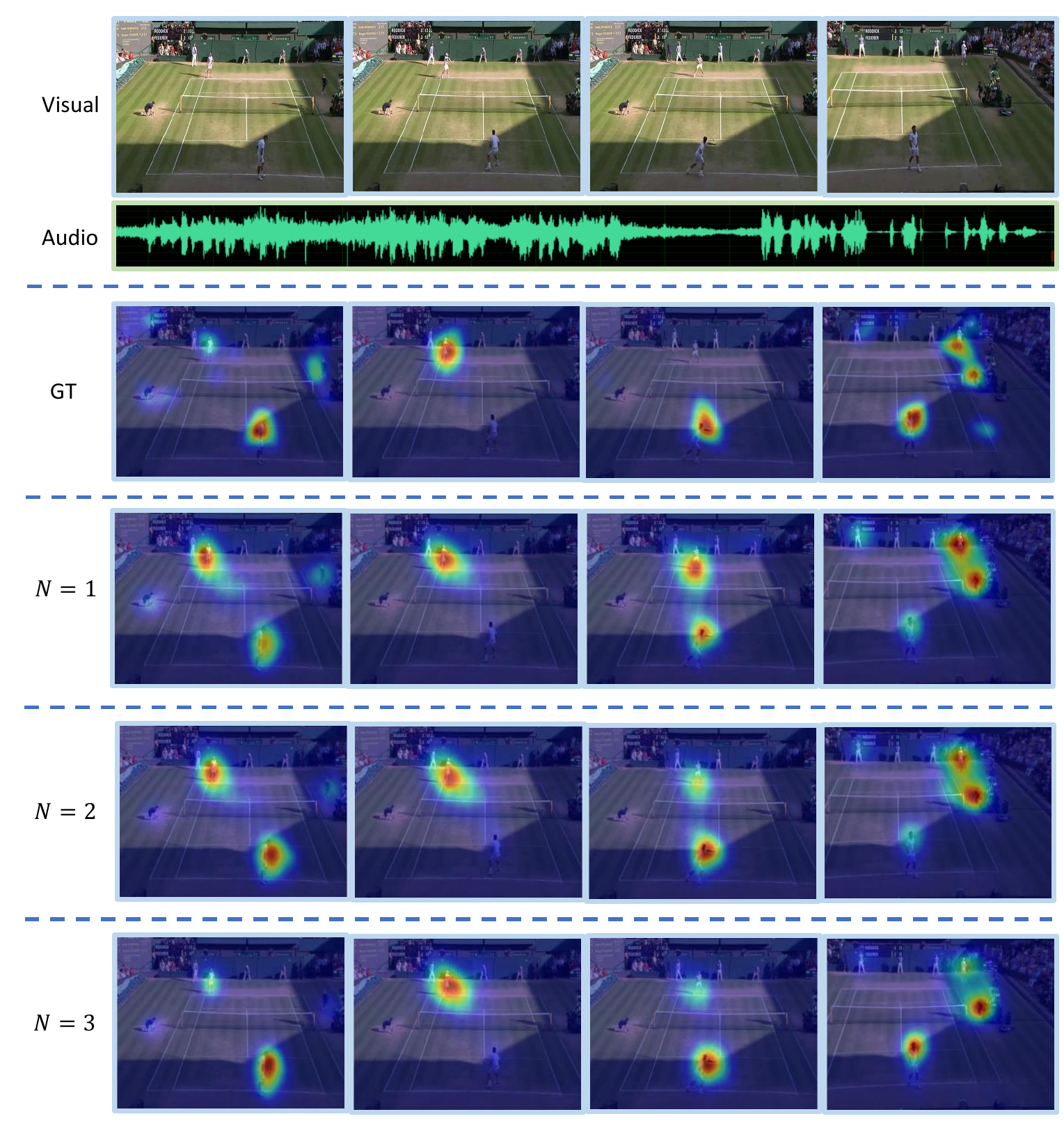}
	\vspace{-10pt}
	\caption{Comparison of saliency maps with different CPC's iterations. 
	 \textit{GT} denotes ground truth.}
	\label{fig-cpc-vis} 
	\vspace{-10pt}
\end{figure}

\begin{table*}[!htbp]
	\begin{center}
		\resizebox{0.95\linewidth}{!}{
			\begin{tabular}{l | c | cccc|cccc|cccc}
				\toprule
				\multirow{2}*{\textbf{Method}} &  \multirow{2}*{\textbf{Pretrained}} & \multicolumn{4}{c}{\textbf{DIEM}} & \multicolumn{4}{c}{\textbf{Coutrot1}} & \multicolumn{4}{c}{\textbf{Coutrot2}}\\
				\cline{3-14}
				& & CC $\uparrow$ & NSS $\uparrow$ & AUC-J $\uparrow$  & SIM  $\uparrow$&  CC $\uparrow$ & NSS $\uparrow$ & AUC-J$\uparrow$   & SIM $\uparrow$ & CC$\uparrow$ & NSS$\uparrow$ & AUC-J$\uparrow$  & SIM$\uparrow$ \\
				\midrule
				\midrule
				ACLNet(V)  \cite{wang2019revisiting}  & - & 0.522 & 2.02 & 0.869 & 0.427 & 0.425 & 1.92 & 0.850  & 0.361 & 0.448 & 3.16 & 0.926 &  0.322 \\
				TASED-Net(V) \cite{min2019tased} & - & 0.557 & 2.16 & 0.881 & 0.461 & 0.479 & 2.18 & 0.867  & 0.388 & 0.437 & 3.17 & 0.921 &  0.314 \\
				STAViS(V)  \cite{tsiami2020stavis} & - & 0.567 & 2.19 & 0.879 & 0.472 & 0.458 & 1.99 & 0.861  & 0.384 & 0.652 & 4.19 & 0.940 & 0.447 \\
				STAViS(AV) \cite{tsiami2020stavis} & - & 0.579 & 2.26 & 0.883 & 0.482 & 0.472 & 2.11 & 0.868  & 0.393 & 0.734 & 5.28 & 0.958 & 0.511 \\
				
				\rowcolor{mygray} CASP-Net(V)  & - & 0.638 & 2.54 & 0.902  & 0.529 & 0.555 & 2.65 & 0.882  & 0.449 & 0.756 & 6.01 & 0.961  & 0.566 \\
				\rowcolor{mygray}CASP-Net(AV) & -  & \textbf{0.649} & \textbf{2.58} & \textbf{0.904}  & \textbf{0.536} & \textbf{0.560} & \textbf{2.66} & \textbf{0.887}  & \textbf{0.453} & \textbf{0.766} & \textbf{6.11} & \textbf{0.963}  &  \textbf{0.573} \\
				
				\hline
				ViNet(V) \cite{jain2021vinet} & DHF1k & 0.626 & 2.47 & 0.898  & 0.483 & 0.551 & 2.68 & 0.886  & 0.423 & 0.724 & 5.61 & 0.95  & 0.466 \\
				ViNet(AV) \cite{jain2021vinet} & DHF1k & 0.632 & 2.53 & 0.899  & 0.498 & 0.56 & 2.73 & 0.889  & 0.425 & 0.754 & 5.95 & 0.951 & 0.493 \\
				TSFP-Net(V) \cite{chang2021temporal}  & DHF1k & 0.649 & \textbf{2.63} & 0.905  & 0.529 & 0.57 & \textbf{2.75} & 0.894  & 0.451 & 0.718 & 5.30 & 0.957  & 0.516 \\
				TSFP-Net(AV) \cite{chang2021temporal} & DHF1k & 0.651 & 2.62 & 0.906  & 0.527 & \textbf{0.571} & 2.73 & \textbf{0.895}  & 0.447 & 0.743 & 5.31 & 0.959  & 0.528 \\
				
				\rowcolor{mygray}CASP-Net(V)  & DHF1k & 0.649 & 2.59 & 0.904  & 0.538 & 0.559 & 2.64 & 0.888  & 0.445 & 0.756 & 6.07 & 0.963 & 0.567 \\
				\rowcolor{mygray}CASP-Net(AV) & DHF1k  & \textbf{0.655} & 2.61 & \textbf{0.906}  & \textbf{0.543} & 0.561 & 2.65 & 0.889  & \textbf{0.456} & \textbf{0.788} & \textbf{6.34} & \textbf{0.963}  &  \textbf{0.585} \\
				\bottomrule
		\end{tabular}}
	\end{center}
	\vspace{-15pt}
	\caption{Comparison of saliency on DIEM, Coutrot1 and Coutrot2 datasets. The experimental table is divided into two groups according to whether the DHF1k dataset is used as pre-training data. We show the modalities used for each method in brackets: (V) for visual, and (AV) for audio-visual.}
	\label{table-sota1}
	\vspace{-5pt}
\end{table*}

\begin{table*}[!htbp]
	\begin{center}
		\resizebox{0.95\linewidth}{!}{
			\begin{tabular}{l |c| cccc|cccc|cccc}
				\toprule
				
				\multirow{2}*{\textbf{Method}}  & \multirow{2}*{\textbf{Pretrained}} & \multicolumn{4}{c}{\textbf{AVAD}} & \multicolumn{4}{c}{\textbf{ETMD}} & \multicolumn{4}{c}{\textbf{SumMe}}\\
				\cline{3-14}
				&  & CC $\uparrow$ & NSS $\uparrow$ & AUC-J $\uparrow$ & SIM  $\uparrow$&  CC$\uparrow$ & NSS$\uparrow$ & AUC-J$\uparrow$  & SIM$\uparrow$ & CC$\uparrow$ & NSS$\uparrow$ & AUC-J $\uparrow$ & SIM$\uparrow$  \\
				\midrule
				\midrule
				ACLNet(V) \cite{wang2019revisiting} & - & 0.580 & 3.17 & 0.905 & 0.446 & 0.477 & 2.36  & 0.915  & 0.329 & 0.379 & 1.79 & 0.868 &  0.296 \\
				TASED-Net(V) \cite{min2019tased} & - & 0.601 & 3.16 & 0.914  & 0.439 & 0.509 & 2.63 & 0.916 & 0.366 & 0.428 & 2.1 & 0.884 & 0.333 \\
				STAViS(V) \cite{tsiami2020stavis} & - & 0.604 & 3.07 & 0.915 & 0.443 & 0.560 & 2.84 & 0.929 & 0.412 & 0.418 & 1.98 & 0.884  & 0.332 \\
				STAViS(AV) \cite{tsiami2020stavis} & - & 0.608 & 3.18 & 0.919 & 0.457 & 0.569 & 2.94 & 0.931 & 0.425 & 0.422 & 2.04 & 0.888  & 0.337 \\

				\rowcolor{mygray} CASP-Net(V) & - &  0.671 & 3.67 & 0.931  & 0.515 & 0.613 & 3.30 & 0.938 &  0.471 & 0.481 & 2.50 & 0.901  & 0.374 \\
				\rowcolor{mygray} CASP-Net(AV) & - & \textbf{0.685} & \textbf{3.77} & \textbf{0.932}  & \textbf{0.528} & \textbf{0.616} & \textbf{3.31} & \textbf{0.939} & \textbf{0.476} &  \textbf{0.486} & \textbf{2.52} & \textbf{0.904} &  \textbf{0.377} \\
				\hline
				
				ViNet(V) \cite{jain2021vinet} & DHF1k & 0.694 & \textbf{3.82} & 0.928 & 0.504 & 0.569 & 3.06 & 0.928 &  0.409 & 0.466 & 2.40 & 0.898  & 0.345 \\
				ViNet(AV) \cite{jain2021vinet} & DHF1k & 0.674 & 3.77 & 0.927 & 0.491 & 0.571 & 3.08 & 0.928 & 0.406 & 0.463 & 2.41 & 0.897  & 0.343 \\
				TSFP-Net(V) \cite{chang2021temporal} & DHF1k & 0.688 & 3.79 & 0.931  & \textbf{0.530} & 0.576 & 3.09 & 0.932  & 0.433 & 0.463 & 2.28 & 0.894  & 0.362 \\
				TSFP-Net(AV) \cite{chang2021temporal} & DHF1k & \textbf{ 0.704 }& 3.77 & 0.932  & 0.521 & 0.576 & 3.07 & 0.932  & 0.428 & 0.464 & 2.30 & 0.894 & 0.360 \\
				
				\rowcolor{mygray} CASP-Net(V) & DHF1k &  0.681 & 3.75 & 0.931 & 0.526 & 0.616 & 3.31 & 0.938  & 0.471 & 0.485 & 2.52 & 0.904 & 0.382 \\
				\rowcolor{mygray} CASP-Net(AV) & DHF1k &  0.691 & 3.81 & \textbf{0.933}  & 0.528 & \textbf{0.620} & \textbf{3.34} & \textbf{0.940}  & \textbf{0.478} & \textbf{0.499} & \textbf{2.60} & \textbf{0.907} &  \textbf{0.387} \\
				\bottomrule
		\end{tabular}}
	\end{center}
	\vspace{-15pt}
	\caption{Comparison of saliency on AVAD, ETMD and SumMe datasets. The experimental table is divided into two groups according to whether the DHF1k dataset is used as pre-training data. 
		We show the modalities used for each method in brackets: (V) for visual, and (AV) for audio-visual.}
	\label{table-sota2}
	\vspace{-15pt}
\end{table*}

\vspace{-15pt}
\subsubsection{Evaluation metrics}
\vspace{-10pt}
For the evaluation of CASP-Net, four widely-used evaluation metrics are adopted \cite{bylinskii2018different}: CC, NSS, AUC-Judd (AUC-J), and SIM. The CC measures the linear correlation coefficient between the ground truth and the predicted saliency map. The NSS focuses on measuring the saliency value on human fixations, the AUC-J is a location-based metric for evaluating the predicted saliency map, and the SIM measures the similarity between the predicted saliency map and the ground truth.


 \vspace{-7pt}
\subsection{Ablation Studies}
Table \ref{table-ablation_1} and Table \ref{table-ablation_2} show ablation studies on different configurations of CASP-Net. All the ablations are performed with the training on the AVAD and ETMD training sets and evaluated on their validation sets. From Table \ref{table-ablation_1}, three different architecture of decoders are compared \wrt their parameters and performance, namely FCN, UNet and Sal (ours). The FCN decoder simply uses  multi-layers network with architecture: 3D Deconv + ReLU + BN. UNet decoder represents the progressive upsampling process that alternates 3D Conv and Trilinear layers while adding UNet skip connections operation. Experimental results show that CASP-Net(Sal) outperforms CASP-Net(FCN) and CASP-Net(UNet) with a smaller parameter quantity. 
It suggests that the decoder FCN and UNet can be replaced in tasks of dense prediction such as saliency prediction, and further confirms the effectiveness of our proposed decoder.

To analyze the performance of each part in the proposed work, a baseline model \textit{Visual-Only} is formed based on a visual encoder and a SalDecoder as shown in Table \ref{table-ablation_2}. Moreover, the Bilinear fusion operation \cite{tenenbaum2000separating} is also introduced to compare with the designed AVIM, the obtained observations are listed below: (i) The combination of AVIM and CPC can continuously improve the performance of the model, which achieves the best performance,
(ii) The linear version of AVIM has roughly the same performance as the quadratic version, but the computational complexity is much lower,
(iii) The Bilinear-based model performs much worse than the AVIM-based model, and even not better than the \textit{Visual-Only} model. This means that the ability of audio-visual feature aggregation from Bilinear operation is inferior to that of the proposed AVIM.

\noindent \textbf{Cross-modal Interaction at Various Stages.} For cross-modal interaction, the AVIM has a plug-in architecture that can be applied in any stage. As shown in the Table \ref{table-AVIM-ablation}, the prediction performance fluctuates when the AVIM is used in different single stage. On both AVAD and ETMD datasets, it is noticed that AVIM performs better in the second and third stages because the acquired semantics of the visual features is limited in the initial stage. Since our SalDecoder  adopts a skip-connection, it would be beneficial to apply the AVIM in multiple stages, as verified in the right part of Table \ref{table-AVIM-ablation}. On both datasets, the best performance is achieved by applying AVIM at four stages, which also indicates the model has the ability to fuse and balance the features from multiple stages.


\noindent \textbf{Further Analysis of CPC.}
The impact of the iterative number of the CPC on its performance also needs to be analyzed. Figure \ref{fig:exp_cpc_iteration} shows that the prediction metrics tend to rise when more iterative computations are performed, especially in the first three iterations. Thus, the iteration number $N$ is set to 3 by default. Figure \ref{fig-cpc-vis} depicts the predicted saliency maps of our method on some audio-visual samples, which shows that the CPC iteratively resolves the internal inconsistencies in audio and visual features. It is also noticed that the saliency maps are inaccurate in early iterations, till later iterations, the model has corrected itself to pay more accurate attention to the objects of interest.

\subsection{Comparisons with State-of-the-art Methods}
\vspace{-5pt}
The proposed CASP-Net is compared with recent state-of-the-art saliency works on six audio-visual datasets as shown in Table \ref{table-sota1} and Table \ref{table-sota2}. The experiment results can be divided into two groups according to whether the DHF1k dataset is used as pre-training data. Results in the two groups highlight the superiority of the proposed audio-visual scheme, as it outperforms the other state-of-the-art works on almost all datasets and metrics. It can be observed that the CASP-Net is able to significantly surpass the prior saliency predictions, such as STAViS \cite{tsiami2020stavis} and TASED-Net \cite{min2019tased}, with the configuration trained on six audio-visual datasets directly. The CASP-Net(AV) achieves an average performance improvement of 11.5\% CC and 13\% SIM compared to STAViS(AV), and becomes a new state-of-the-art on six benchmarks.



The proposed work also shows an obvious superiority in audio-visual saliency prediction with the DHF1k pre-training dataset. Compared to ViNet \cite{jain2021vinet} and TSFP-Net \cite{chang2021temporal}, the CASP-Net(AV) achieves the best performance in most datasets, especially on the Countrot2, ETMD and SumMe test sets. Moreover, after incorporating audio features into the pure visual model, the relative improvement on these three datasets is the most significant (average 3 \% relative improvement in CC  and 2 \% relative improvement in SIM). This reflects the high audio-visual correspondence of three datasets, as well as the model's capability to take full advantage of these audio-visual cues. Overall, the experiment results in both groups have shown that pre-training on a large-scale video dataset enables the proposed CASP-Net to be effectively generalized on the other datasets. For qualitative analysis, the CASP-Net is further compared with the STAViS and ViNet on Coutrot2, ETMD, AVAD and SumMe datasets to show superior performance.

\begin{figure}[!htbp]
	\includegraphics[scale=0.48]{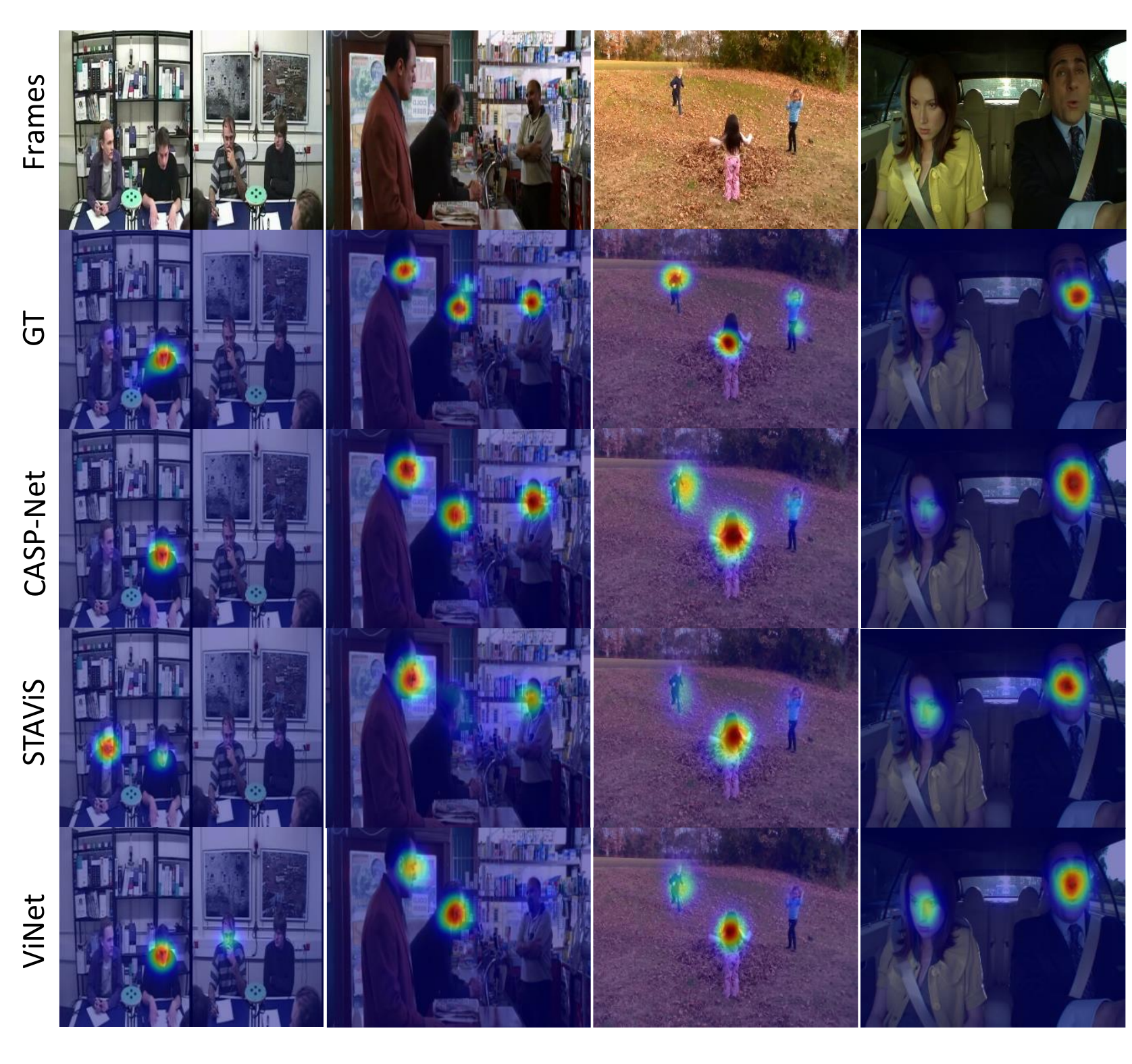}
	\vspace{-10pt}
	\caption{Sample frame from Coutrot2, ETMD, AVAD and SumMe databases with their eye-tracking data, and the corresponding ground truth, CASP-Net, and other state-of-the-art audio-visual saliency maps for comparisons. }
	\label{fig-multi_speaker}
\end{figure}

In Figure \ref{fig-multi_speaker}, the advantages of the proposed work have been demonstrated in the audio-visual scenarios having multiple speakers. The first and second rows are the frames and the corresponding ground truth saliency map. The third row presents the prediction saliency maps from our CASP-Net, and the final 2 rows are the same maps for STAViS and ViNet. In particular, our results are closer to the ground truth.

\vspace{-10pt}

 \vspace{-25pt}
\section{Conclusion}
 \vspace{-5pt}
We propose a consistency-aware audio-visual saliency prediction network (CASP-Net), which  effectively addresses potential audio-visual inconsistency in video saliency prediction. 
A two-stream network  is designed to elegantly associate video frames  with the corresponding sound source, achieving cross-modal semantic similarities between audio and visual features. In addition, a novel consistency-aware predictive coding module is introduced to improve the consistency within audio and visual representations iteratively. Besides, a saliency decoder is also designed to aggregate the multi-scale audio-visual information and obtain the final saliency map. Experiments show surprising results that CASP-Net outperforms 5 state-of-the-art approaches on 6 challenging audio-visual eye-tracking datasets.

\noindent \textbf{Acknowledgements.} This work was supported by NSFC under Grants 61971352, 62271239, 61862043, Ningbo Natural Science Foundation  under Grants 2021J048, 2021J049.

{\small
\bibliographystyle{ieee_fullname}
\bibliography{egbib}
}

\newpage
\appendix

\end{document}